\documentclass[conference]{IEEEtran}  

\usepackage{graphicx}
\usepackage{float}
\usepackage{tikz}
\usepackage{amsmath,amsthm}
\usepackage{amssymb}
\usepackage{algpseudocode}
\usepackage{epstopdf}

\usepackage{multirow}

\usepackage{subfig}
\usepackage{balance}
\usepackage{color}

\newtheorem{stp}{Step}

\usepackage{bbm}

\title{\LARGE \bf Towards Controllability of Wireless Network Quality using Mobile Robotic Routers}

\author{
\IEEEauthorblockN{Pradipta Ghosh\IEEEauthorrefmark{1}, Raktim Pal\IEEEauthorrefmark{2} and Bhaskar Krishnamachari\IEEEauthorrefmark{1}}
\IEEEauthorblockA{\IEEEauthorrefmark{1}Ming Hsieh Department of Electrical Engineering, University of Southern California \{pradiptg, bkrishna\}@usc.edu}
\IEEEauthorblockA{\IEEEauthorrefmark{2}Department of Electrical and Computer Engineering, Georgia Institute of Technology \{raktimnitt@gmail.com\}}
}

\begin{document}

\maketitle
\thispagestyle{empty}
\pagestyle{empty}

\begin{abstract}
We consider a problem of robotic router placement and mobility control with the objective of formation and maintenance of an optimal communication network between a set of transmitter-receiver pairs.
In this scenario, the communication path between any transmitter-receiver pair contains a predetermined set of mobile robotic routers nodes.
The goal of this work is to design an algorithm to optimize the positions of the robotic nodes to improve the overall performance of the network. 
We define the optimization metric to be the minimum of the Signal to Interference plus Noise Ratios (SINR) over all the links.
In this manuscript, we propose two optimization algorithms to solve this problem in a centralized and a decentralized manner, respectively.
We also demonstrate the performances of both algorithms based on a set of simulation experiments.
\end{abstract}


\section{Introduction}

Distributed cooperation in mobile robotics is a very important domain of research that mainly focuses on motion and position configurations of a group of robots to perform a set of tasks.
To this end, researchers have proposed a range of algorithms based on information exchanges between robots such as swarming\ \cite{gazi2002stability}, flocking \cite{olfati2006flocking} and formation control~\cite{fax2004information}.
The applications of cooperative robotics range across different domains 
such as search and rescue operations, underground mining, remote explorations, fire-fighting and military operations. 
However, effective cooperation between robots in any such application depends on the capacity and reliability of the wireless communication infrastructure.
Conversely, a group of cooperative robots can be utilized to improve the performance, capacity and reliability of
a wireless communication infrastructure and even to build a controllable wireless communication backbone.
For example, a group of robots with wireless communication and routing\
capabilities can form a communication path between a sender and a receiver\
that are unable to directly communicate with each other.\
In summary, cooperative robotics and wireless networks are two symbiotic fields of research.

Till last decade, there was no significant research on the applications of cooperative robotics towards the advancement and improvement of traditional communication infrastructures.
To this end, research on robotic routers and relay agents in sensing and information routing is an emerging research domain.
An important and fundamental problem in this domain is to guarantee an optimal, efficient and fair flow of information in the network.
Note that the definitions of optimality and fairness themselves depend on the application scenarios.
For example, optimality can be defined in terms of signal to interference and noise ratio (SINR), proper allocation of robots between different flows, or bit error rates (BER).
Another challenge is to improve the overall performance and quality of a network in a distributed manner instead of a centralized way.
All these problems are directly related to the proper placement of robotic routers.
Therefore, we primarily focus on optimizing the robotic router placements in order to optimize the overall network performance.

The goal of this work is to devise an algorithm for proper placement and movement control of the robotic router nodes.
We follow three main steps towards achieving this goal.
\emph{First,} we identify the advantages and disadvantages of different optimization metrics to achieve the goals.
Some of the potential choices are the expected transmission count metric (ETX), the bit error rate (BER) and the signal-to-interference-plus-noise ratio (SINR).
However, we choose SINR as the optimization metric because it is directly related to the communication quality and because the rest of the metrics can be expressed as a function of SINR.
\emph{Second,} we design a proper optimization function that is directly related to the optimization goals.
\emph{Third,} we propose two optimization techniques to efficiently solve the optimization problem. 
The first technique is a centralized algorithm which is based on a well known stochastic optimization technique called Simulated Annealing (\cite{van1987simulated,aarts1988simulated}).
The second one is a distributed technique where each node makes movement control decisions in a distributed manner in order to improve the overall quality of the network.
This technique is subdivided into two major parts which we refer to as \textbf{S}INR Da\textbf{T}a \textbf{A}ccumulation \textbf{T}ask (STAT) and \textbf{Mo}vement \textbf{D}irection and Gr\textbf{A}nularity Contro\textbf{L} (MODAL), respectively.
We also present a simulation based performance and correctness analysis of our proposed methods.




\section{Problem Formulation}
\label{sec:problem}
In this section, we discuss our problem formulation in details. 
First of all, for compactness, we summarize the symbols used for this formulation in Table~\ref{tab:symbol}.
We assume that each node of the network has a unique ID as well as proper localization techniques.
The set of node locations is represented as ${X}=\{ (x_i,y_i): i \mbox{ is the node ID}\}$.
We consider a network with $n$ transmitter-receiver pairs, which we also refer to as the communication endpoints to avoid ambiguity, denoted by $ TX_i=\{ T_i, R_i \}$ where $T_i$ and $R_i$ are the IDs of $i$th sender and receiver, respectively.
For each transmitter-receiver pair, say $i$th pair, we associate a set of robots, $\mathcal{M}_i$.
The total number of robots in the system is $m$, i.e., $\sum_{i=1}^n |\mathcal{M}_i| = m$. 
We define a `flow' to be the set of links that form the communication path between a transmitter and its corresponding receiver.
For example, $L_i=\{ L_{ij}: j=1,\cdots,|\mathcal{M}_i|+1 \}$ represents the $i^{th}$ flow.
\begin{table}[]
    \caption{Symbol Table}
    \centering
    \begin{tabular}{|c|c|}
    \hline
    Symbol & Description \\ \hline
       $X$  & Set of Node locations \\ 
       $X_{\mathcal{M}}$   & Set of Robots' locations \\
       $T_i$  & Transmitter of Flow $i$\\
       $R_i$ & Receiver of Flow $i$\\
       $TX_i$ & Communication Endpoints for flow $i$ \\
       &i.e., $\{ T_i, R_i \}$\\
      $\mathcal{M}_i$ & Set of robots allocated to flow $i$ \\
      $ L_{ij}$ & $j^{th}$ link of flow $i$, numbered from $T_i$\\
      $P\left(L_{ij}\right)$ & Transmission Power for link $L_{ij}$\\
      $d\left(L_{ij}\right)$ & Length of link $L_{ij}$\\
      $\eta$ & Path loss exponent  \\
      $P_{inter,k}\left(L_{ij}\right)$ & Transmission power of the link $L_{ij}$'s\\
      & $k^{th}$ interference source \\
      $d_{inter,k}\left(L_{ij}\right)$ &  Distance between the link $L_{ij}$'s receiver end\\
      & and its $k^{th}$ interference source \\
      $\psi_{dB},\psi^k_{dB}$ &   Log normal fading effect $\sim \mathcal{N}(0,\sigma^2)$\\
      $P_N\left(L_{ij}\right)$ &   Noise power for the link $L_{ij}$ \\
      \hline
    \end{tabular}
    \label{tab:symbol}
\end{table}

We select SINR as the optimizing quantity.
The  overall performance of a network depends on the SINR quality of each of the links.
The link with the minimum SINR restricts the overall performance of a network and acts as the bottleneck.
Therefore, an important step towards optimizing a network is to try to improve the SINRs of each and every link.
However, SINRs of most of the links are not independent, thereby making the optimization task complex.
Ideally, we need an optimization function that improves the overall performance of a network while considering the link dependencies.
In this context, our optimization goal is to maximize the minimum SINR of a network.
We define the cost function for each flow, say $\{i\}$, to be as follows. 

{\footnotesize
\begin{equation}
\mathcal{C}_i(X_\mathcal{M})=\underset{j}{\min } \  SINR\left(L_{ij},X_\mathcal{M}\right)
\label{eq:cost1}  
\end{equation}
}
where $X_\mathcal{M} \subset X$ is the set of all robots' locations and $SINR\left(L_{ij},X_\mathcal{M}\right)$ denotes the SINR of link $L_{ij}$ for the configuration $X_\mathcal{M}$.
Therefore, the overall cost function for the entire network is 
{\footnotesize
\begin{equation}
\mathcal{C}(X_\mathcal{M})=\underset{i}{\min } \  \mathcal{C}_i(X_\mathcal{M})
\label{eq:cost}  
\end{equation}
}
Now, the optimization goal is as follows.
{\footnotesize
\begin{equation}
\underset{X_\mathcal{M}}{maximize}\{\mathcal{C}(X_\mathcal{M})\}
\label{eq:opt}  
\end{equation}
}
Using simple path loss model and log-normal fading model~\cite{rappaport1996wireless}, the SINR of a link, $L_{ij}$, can be mathematically represented as follows.

{\footnotesize
\begin{align}
\label{eqn:sinr}
\frac{P\left(L_{ij}\right) d\left(L_{ij}\right)^{-\eta}10^{ \frac{\psi_{dB}}{10}}}{\underset{k}{\sum} P_{inter,k}\left(L_{ij}\right)\
 d_{inter,k}\left(L_{ij}\right)^{-\eta}10^{ \frac{\psi^k_{dB}}{10}} + P_N\left(L_{ij}\right)}
\end{align}
}
where the meaning of each symbol is illustrated in Table~\ref{tab:symbol}.
We simplify this problem by assuming that all robot's transmission power are identical, say $P_M$, all sender's transmission power are identical, say $P_T$, and the noise powers are constant, say $P_N$.
In this work, we are mainly interested in demonstrating a proof of concept of robotic router placement with the goal of SINR optimization. Thus, we further simplify the model by ignoring the fading effect i.e. taking $\psi_{dB}=0,\psi^k_{dB}=0$ to deal with only the mean power value. Including the fading effect will only delay the convergence of the process.  
Our algorithms work for any value of $\eta$.
However, all the experiments presented in this paper are performed with $\eta =2 $.
\textbf{Note that we do not assume the communication endpoints to be static i.e., the endpoint can be mobile.}

\subsection{Sample Problem:}
\begin{figure}[ht!]
\centering
\includegraphics[width=.6\linewidth]{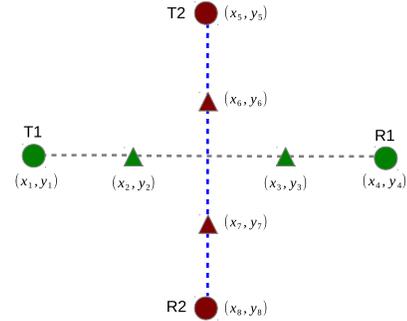}
\caption{A Simple Example Scenario}
\label{fig:2pair}
\end{figure}

Let's consider a very simple configuration with only two pairs of transmitter-receiver.\
We deploy four robotic nodes, which facilitate communication between them, with\
two robots for each flow as in Figure~\ref{fig:2pair}.\
There are total 8 nodes with IDs: $1$ to $8$ respectively.
In this example, $ TX_1=\{ T_1, R_1 \}=\{ 1, 4 \}$, $TX_2=\{ T_2, R_2 \}=\{ 5, 8 \}$,\
$\mathcal{M}_1=\{ 2, 3 \}$ and $\mathcal{M}_2=\{ 6, 7 \}$.
The transmitters and the receivers are assumed to be static with coordinates $\{x_i, y_i\}$ for $ i = 1, 4, 5, 8$\
and the robotic nodes are mobile with coordinates $\{x_i, y_i\}$ for $ i = 2, 3, 6, 7$.\


In Table~\ref{table:inter}, we present a list of interfering nodes for each link,\
which is essential to calculate the SINRs.\
\begin {table}[H]
\small
\caption{Interference Table}
\label{table:inter}
\centering
\begin{tabular}{| c | c ||c|c|}
\hline
Link ID & Interfering & Link ID & Interfering\\
&Node IDs&&Node IDs\\
\hline
$L_{12}$ & 3, 5, 6, 7& $L_{56}$ & 1, 2, 3, 7\\ \hline
$L_{23}$ & 1, 5, 6, 7 & $L_{67}$ & 1, 2, 3, 5\\ \hline
$L_{34}$ & 1, 2, 5, 6, 7& $L_{78}$ & 1, 2, 3, 5, 6 \\ \hline
\end{tabular}
\end {table}
The SINRs at node 2, 3 and 4 are represented as: $SINR(L_{12})$, $SINR(L_{23})$, $SINR(L_{34})$, respectively. 
 {\footnotesize
 \begin{equation}
 SINR(L_{12})=\frac{P_T d_{12}^{-\eta}}{P_T d_{52}^{-\eta} + P_M\left(d_{32}^{-\eta}+d_{62}^{-\eta}+d_{72}^{-\eta}\right)+P_N}
 \end{equation}
 \begin{equation}
 SINR(L_{23})=\frac{P_M d_{23}^{-\eta}}{P_T \left(d_{13}^{-\eta}+d_{53}^{-\eta}\right) + P_M\left(d_{63}^{-\eta}+d_{73}^{-\eta}\right)+P_N}
 \end{equation}
 \begin{equation}
 SINR(L_{34})=\\
 \frac{P_M d_{34}^{-\eta}}{P_T \left(d_{14}^{-\eta}+d_{54}^{-\eta}\right) + P_M\left(d_{24}^{-\eta}+d_{64}^{-\eta}\
 +d_{74}^{-\eta}\right)+P_N}
 \end{equation}
 }
where $d_{ij}$ denotes the euclidean distance between node i and node j.\ 
Similarly, we can write the SINR at nodes 6, 7 and 8 are represented as $SINR(L_{56}), SINR(L_{67}), SINR(L_{78})$, respectively.
We assume that there exists no collision avoidance mechanism\cite{bianchi2000performance} to avoid interference. \emph{ While we acknowledge that this assumption is unreasonable for real systems, we argue that if our algorithm can handle the worst possible case of interference (i.e., without any collision avoidance mechanism), it will also be able to work well in CSMA based systems or equivalent systems.}
Now, the optimization goal is to maximize the minimum of these six SINR values.

\subsection{Properties of the Optimization Function:}
\label{sec:soln}
\begin{figure}[ht!]
\centering
\includegraphics[width=.6\linewidth]{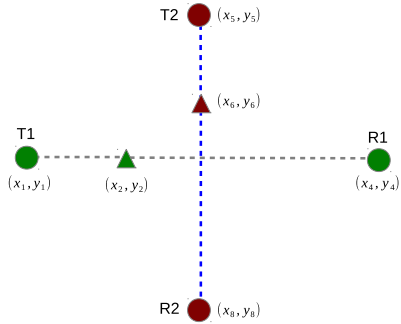}
\caption{Simplified Problem}
\label{fig:property}
\end{figure}

\begin{figure}[ht!]
\centering
\includegraphics[width=.9\linewidth]{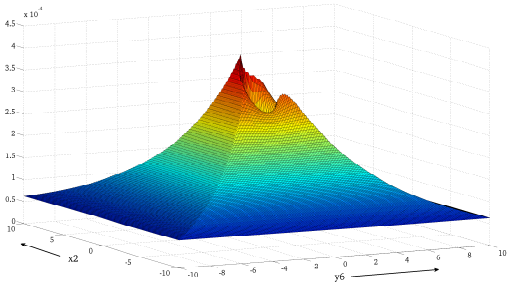}
\caption{Plot of minimum SINR}
\label{fig:3dplot}
\end{figure}

Similar to any optimization problem, it is important to understand\
the behavior of our optimization function in order to identify suitable optimization tools and techniques.\
To analyze the optimization function proposed in~\eqref{eq:opt},\
we take a scenario with two flows, where only one mobile node is allocated to each\
flow as in Figure~\ref{fig:property}.\
In this process, we move the robotic routers along the straight line between the sender\
and the receiver of the respective flow.\
We plot the minimum SINR in the network as a function of the coordinates of these\
two mobile nodes in Figure~\ref{fig:3dplot}.\
From the figure, it is clear that the optimization function is neither convex nor concave.\
It has two peaks with \textbf{different performance}, corresponding to two unique sets of robotic nodes' positions.
From this observation, we conclude that the original generalized problem is non-convex.\
Henceforth, we can not use traditional convex optimization algorithms.\
Note that, the optimization problem may be convex under some special circumstances.\

\section{A Centralized Solution}
\label{sec:central}
In this section, we present a centralized method of solving this problem.
While centralized ways are much easier and straightforward, they are less practical than decentralized approaches.

\subsection{Simulated Annealing based approach}
In this method, we assume that there exists either a central server or a leader node that can communicate with all nodes in the system.
The central server has online knowledge of the positions of all the nodes, ${X}$, and SINRs of all the links in the network.
The central server can either calculate the SINRs of all links using proper signal strength model or directly collect the SINR measurements from each individual nodes.
However, in this paper, we use simple path loss model~\cite{rappaport1996wireless} for simplicity, while more realistic signal strength modeling is left as a future work.
For the optimization purpose, we use a well-known stochastic global optimization algorithm called Simulated Annealing (\cite{van1987simulated,aarts1988simulated}) which can be used to find out the global optimum of complex problems with a large search space.
At each step of this algorithm, a new set ${X'_\mathcal{M}}$ is generated. 
However, each new point $\left(x',y'\right)$ should be in the neighborhood of the original point\
$\left(x,y\right)$, i.e., $ \left(x',y'\right) \in \ \mathcal{N}\left(x,y\right)$,\
where $\mathcal{N}(\cdotp, \cdotp)$ refers to the neighborhood of a location.\
If the set, ${X'_\mathcal{M}}$, has a higher cost function than $X_\mathcal{M}$, the new set ${X'_\mathcal{M}}$ is accepted unconditionally.\
In other words, if $\mathcal{C}({X'_\mathcal{M}})>\mathcal{C}({X_\mathcal{M}})$, new ${X_\mathcal{M}}={X'_\mathcal{M}}$.
However, if $\mathcal{C}({X'_\mathcal{M}})\leq \mathcal{C}({X_\mathcal{M}})$, $X'_\mathcal{M}$ is accepted\
probabilistically using the Metropolis criterion.\
According to the Metropolis criterion, the probability of $X'_\mathcal{M}$ being selected is 

{\footnotesize
\begin{align}
 p = min\left(1,exp\left[-\frac{\mathcal{C}\left( {X_\mathcal{M}} \right)-\mathcal{C}\left( {X'_\mathcal{M}}\right)}{T}\right]\right)
\end{align}
}

Initially, when $T$ is high, there is a greater probability of making downhill moves,\
which allows the algorithm to fully explore the space.
We choose the proper annealing schedule and the number of iterations based on a number of simulations\
and by taking into account the percentage of uphill moves versus the temperature.\
\begin{figure}[ht!]
\centering
 \subfloat[]{\label{fig:f1}\includegraphics[width=0.45\linewidth]{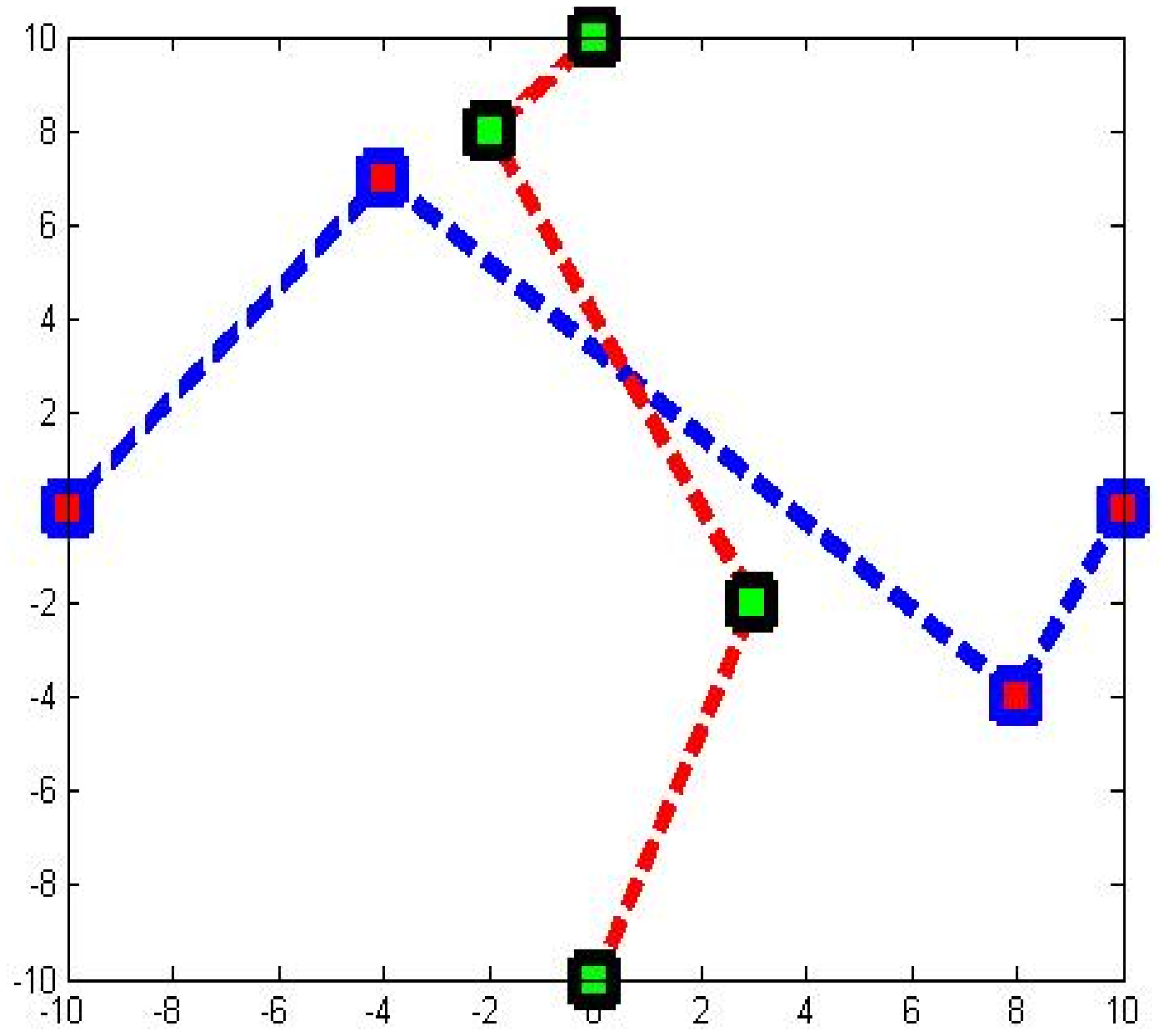}}
 \subfloat[]{\label{fig:f25}\includegraphics[width=0.45\linewidth]{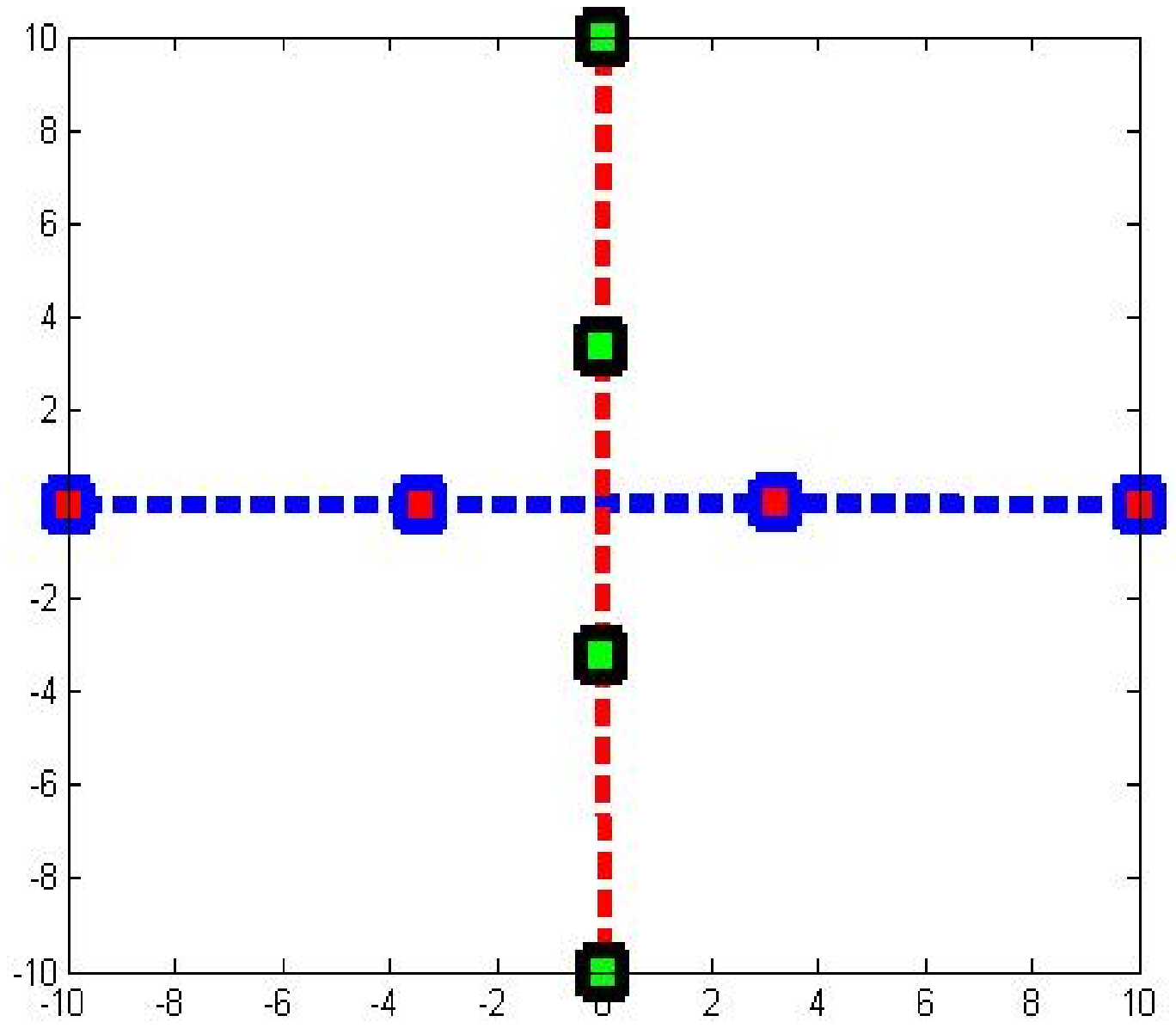}} 
\caption{{Simulation plots for Simulated Annealing (a) Initial Configuration\
 (b) Configuration after 100 Iterations}}
\label{overflow1}
\end{figure}

\subsection{Simulation}
\label{sec:simul1}
We performed a set of simulations on MATLAB 8.1, on a machine with 3.40 GHz Intel i7 processor and 12GB RAM\
to check the convergence of the algorithm for different initial configurations.\
For this experiment, we considered the topology introduced in Figure~\ref{fig:2pair} with the transmitters fixed\
at co-ordinates $(-10,0)$ and $(0,10)$ and the respective receivers fixed at $(10,0)$ and $(0,-10)$.\
We observed that the Simulated Annealing algorithm converges to the same final configuration irrespective\
of different initial configurations of the robots.\
Table~\ref{table:nonlin} illustrates the final SINR values of different links of the network\
for different initial configurations and noise values.\
The initial and final configurations of the robotic nodes for one of the simulation\
instances is presented in Figure~\ref{overflow1}.\
It is clear from the simulation results that the SINR values of all the links are\
equal after the optimization process, which is\
quite intuitive from the symmetry in the network structure.\
The network cannot be further improved in terms of overall performance.


\begin{table*}
{
\caption{SINR values of the different links after Simulated Annealing}
\label{table:nonlin}
\centering
\hfill{}
\begin{tabular}{|c|c|c|c|c||c|c|c|c|c|c|}
\hline
\multicolumn{5}{|c||}{Initial Configuration of Robots} & \multicolumn{6}{|c|}{Final SINR on each link}\\ \hline
$(x_2, y_2)$ & $(x_3, y_3)$ & $(x_6, y_6)$ & $(x_7, y_7)$ & Noise & Link 12 & Link 23 & Link 34 & Link 56 & Link 67 & Link  78\\[1ex]
\hline
(0, 0) & (0, 2) & (0, 0) & (0.5, 0) & 0.6 & 0.0327 & 0.0327 &  0.0327 & 0.0327 & 0.0327 & 0.0327\\[0.5ex]
(0, 0) & (2, 0) & (0, 0) & (0, 0) & 1.0 &  0.0200 & 0.0200 &  0.0200 & 0.0200 & 0.0200 & 0.0200\\[0.5ex]
(0, -1) & (0, 1 ) & (0, 0) & (0, 0) & 2.0 &  0.0108 &  0.0108 & 0.0108 & 0.0108 & 0.0108 & 0.0108\\[0.5ex]
(0, 0) & (3, 0) & (0, 1) & (-1, 0) & 3.0 &  0.0073 &  0.0073 & 0.0073 & 0.0073 & 0.0073 & 0.0073\\[0.5ex]
(0, 2) & (0, 0) & (0, 0) & (-1, 0) & 4.0 &  0.0055 & 0.0055 & 0.0055 & 0.0055 & 0.0055 & 0.0055\\[0.5ex]
(0, 0) & (0, 0) & (0, 0) & (0, 0) & 10.0 &  0.0022 & 0.0022 & 0.0022 & 0.0022 & 0.0022 & 0.0022\\[0.5ex]
\hline
\end{tabular}}
\hfill{}

\end{table*}

\section{Distributed Optimization}
\label{sec:distributed}
In this section, we propose a new distributed approach for solving the optimization problem.
In this distributed approach, each mobile node makes local decisions based on SINR measurements and moves according to those decisions in order to improve the overall quality of the network.

\begin{stp}
\label{stp:1}
Each node calculates the SINRs of its incoming links and communicates these locally calculated SINR values with all the other nodes that belongs to the same flow, whenever a SINR is updated.
\end{stp}
We assume that every node have the necessary hardware and techniques to calculate the SINRs.
\begin{stp}
\label{stp:2}
Each node utilizes the gathered SINR information to determine whether it is a part of the link that has the lowest SINR (which we refer to as the bottleneck link) or the second lowest SINR (which we refer to as the pseudo-bottleneck link) for the respective flow.
\end{stp}
We refer to the Steps~\ref{stp:1} and~\ref{stp:2} together as the \textbf{S}INR Da\textbf{T}a \textbf{A}ccumulation \textbf{T}ask (STAT).
\begin{stp}
If a robotic node $v$ is a part of the bottleneck link (or the pseudo-bottleneck link) of a flow $i$, it makes a local control decision about its movement and moves accordingly, in order to improve the link's SINR, if an improvement is possible without worsening the flow cost i.e., $\mathcal{C}_i(X_\mathcal{M})$.
\end{stp}
We consider the second lowest SINR link to add some diversity in our algorithm so that it doesn't get stuck when no improvement is possible by just moving the bottleneck link's mobile endpoints.
We refer to this step as the \textbf{Mo}vement \textbf{D}irection and Gr\textbf{A}nularity Contro\textbf{L} (MODAL).\
In this step, each node decides the movement direction and granularity as follows.\

We discretize the movements into steps of $\delta > 0$, which needs to be carefully chosen.\
The value of $\delta$ can be adapted based on past movement history in order to increase the speed of convergence.\
However, in this paper, we use a constant value of $\delta$ for simplicity.
Therefore, at each step, a robot $\{v\}$ can move to any point of the circumference of a circle with radius $\delta $ centered at the robot's current location. 
We denote this circle as $\mathcal{B}_{\delta}\{\mathbf{x}_{v}\}$, where $\mathbf{x}_{v}=(x_v,y_v)$ is the current location of the robot $\{v\}$.
To determine the best direction of movement, each robot needs information about link qualities at every possible future locations.
Now, assume that each robot have necessary SINR information about all potential future locations.
Then, a robot $\{v\}$ simply employs a potential based controller for the movements by setting the potential of each future location, say $\mathbf{x'}_{v}=(x'_v,y'_v)$, to be negative of the cost of the flow for the new configuration, $\mathcal{C}_i(X'_\mathcal{M})$ where $X'_\mathcal{M}=\{X_\mathcal{M}\setminus \mathbf{x}_{v}\}\cup \mathbf{x'}_{v}$ and $i$ is the flow that the robot is part of. 
Therefore, the gradient of the potential will determine the best direction.
To gather information about SINR in future locations, we propose two different strategies.
First, we select a finite set of uniformly distributed points, say $\mathbf{x}^f_v$, from the set of possible future locations.
In our case, the set of possible future locations is the circumference of the circle $\mathcal{B}_{\delta}\{\mathbf{x}_{v}\}$. 
Thus we choose a set of points, say 36 points, over the circumference that are equidistant. 
Now, in the first proposed strategy, a robot simply moves to each of these points $\mathbf{x}\in \mathbf{x}^f_v$ and calculates the SINR, assuming that the rest of the network is unchanged. 
Although this method is straightforward, the convergence time of this method is very long as each iteration needs a significant amount of time and it is not efficient in terms of energy consumption.
In the second method, each robot maintains a SINR belief model of the network and updates it after every movement.
Based on that model, a robot estimates the SINRs for each of the potential locations. 
However, this SINR belief model is part of our ongoing work.
\begin{figure}[!ht]
 \centering
 \subfloat[]{\label{fig:f1}\includegraphics[width=0.45\linewidth]{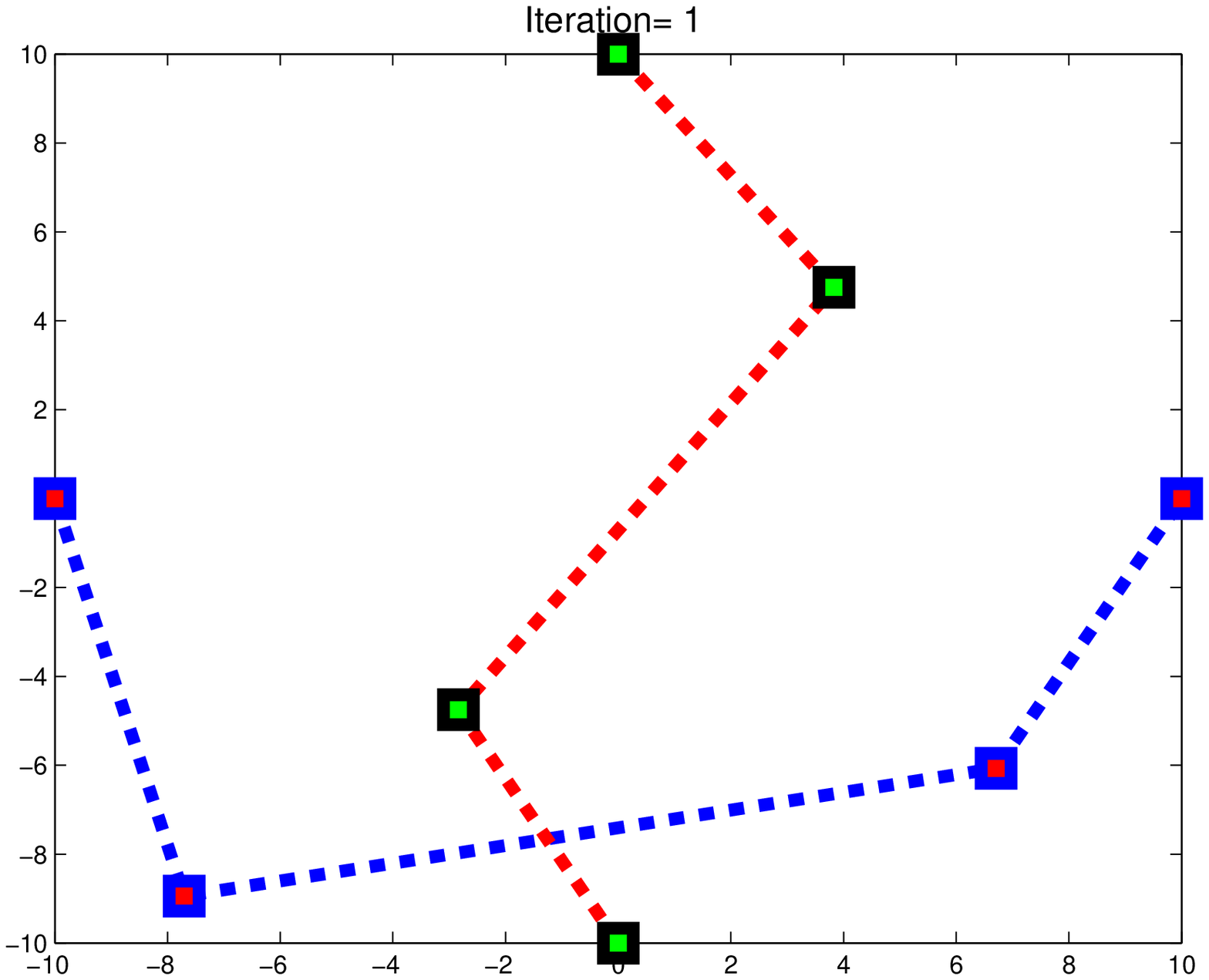}}
 \subfloat[]{\label{fig:f100}\includegraphics[width=0.45\linewidth]{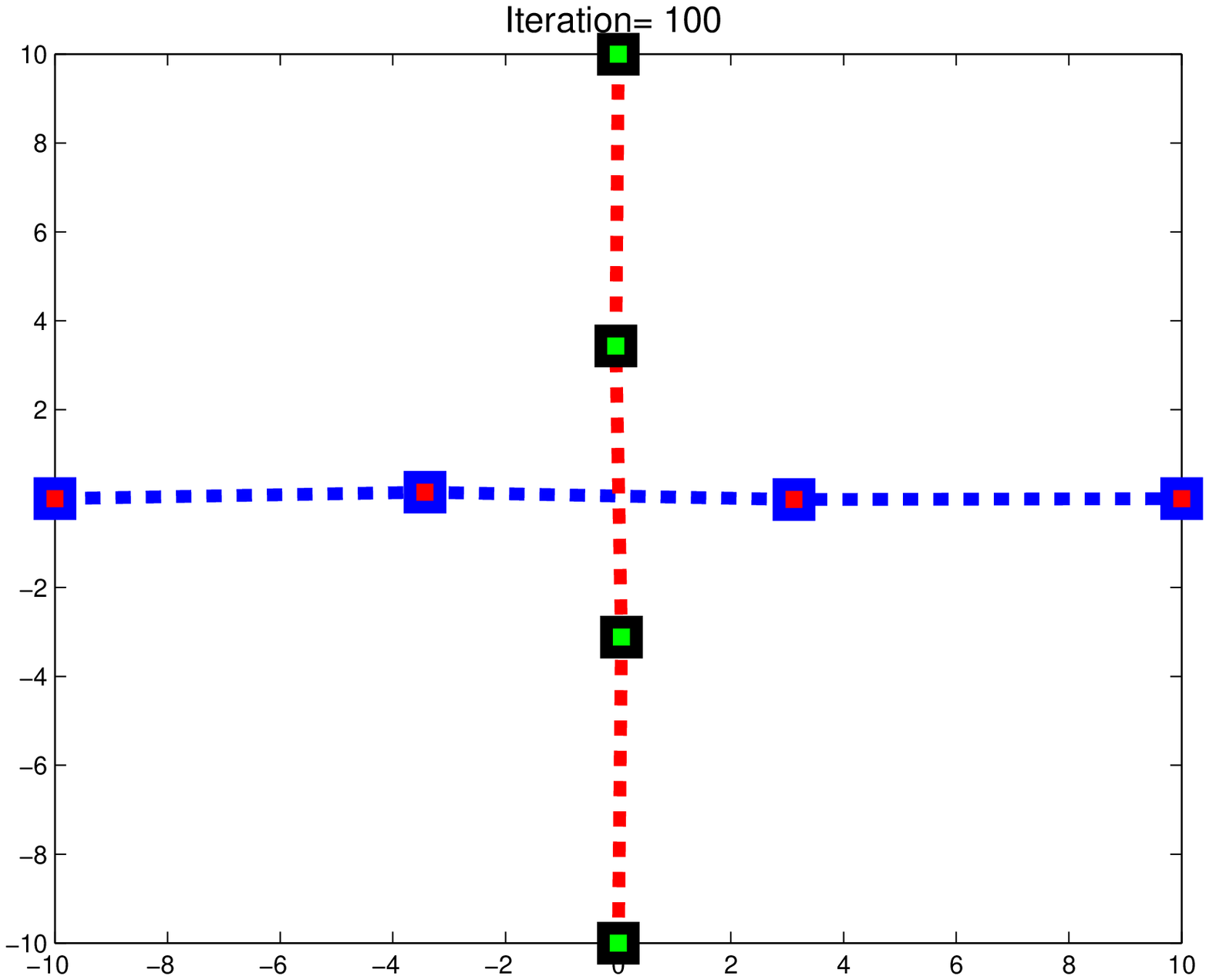}} 
 \caption{{Simulation plots (a) Initial Configuration\
 (b) Configuration after 100 Iterations}}
 \label{fig:distributed}
\end{figure}
Both parts of our algorithm i.e., STAT and MODAL, are repeated until further improvement is possible for neither the link with lowest SINR nor the link with second lowest SINR. The process will restart if any robot senses a change in the configuration.

\subsection{Simulation}
To check the performance and determine the properties of this algorithm, we performed a set of simulation experiments with the same initial configuration as described in section~\ref{sec:simul1}.
One instance of such experiments is illustrated in Figure~\ref{fig:distributed}.
As the figure suggests, the result is strikingly similar to the one obtained from the centralized approach.
Also, with the different initial configuration as in Table~\ref{table:nonlin}, the final SINRs are exactly same as in Table~\ref{table:nonlin}.
\begin{figure}[!ht]
 \centering
 \subfloat[]{\label{fig4:f1}\includegraphics[width=0.45\linewidth]{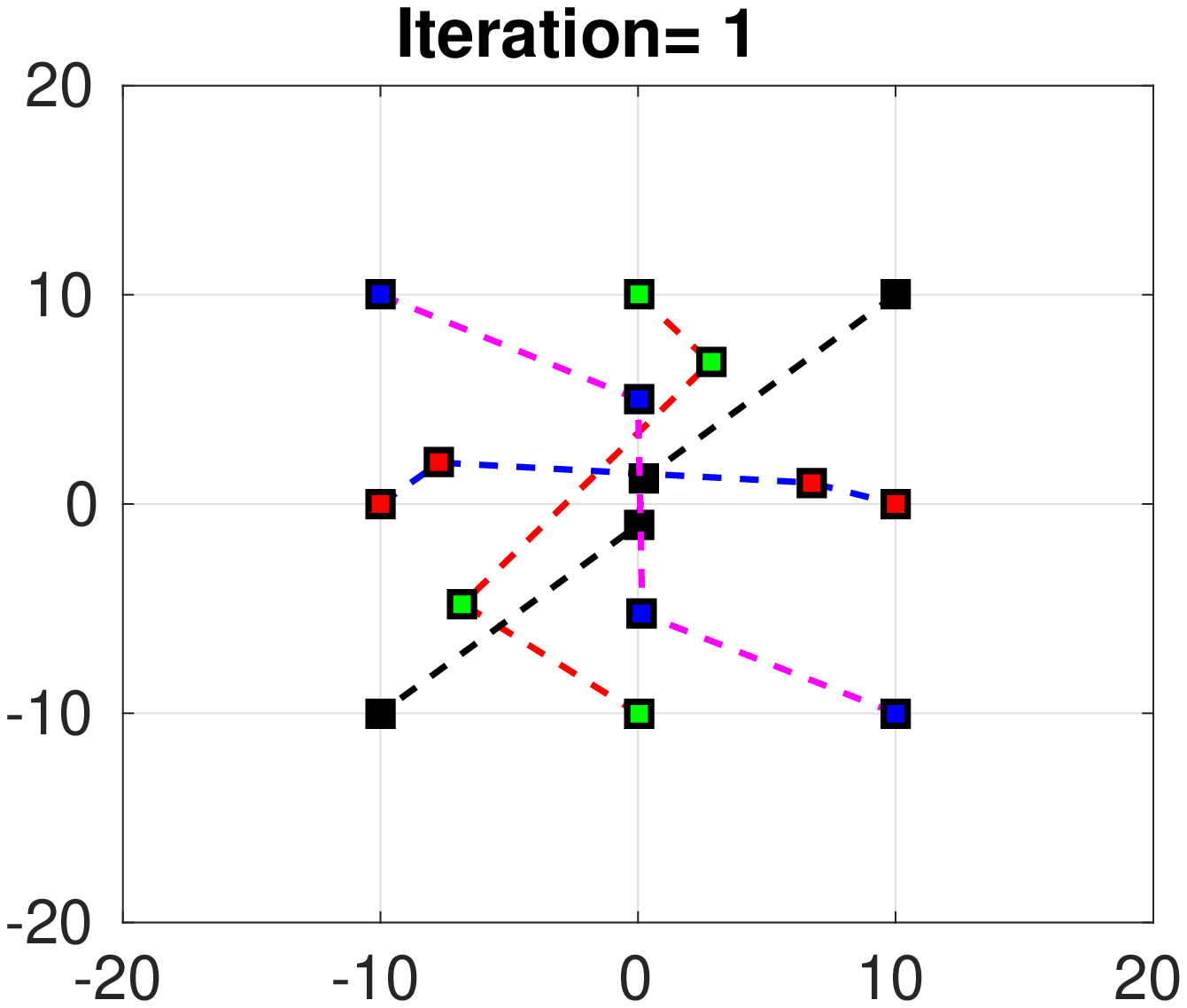}} \ 
 \subfloat[]{\label{fig4:f75}\includegraphics[width=0.45\linewidth]{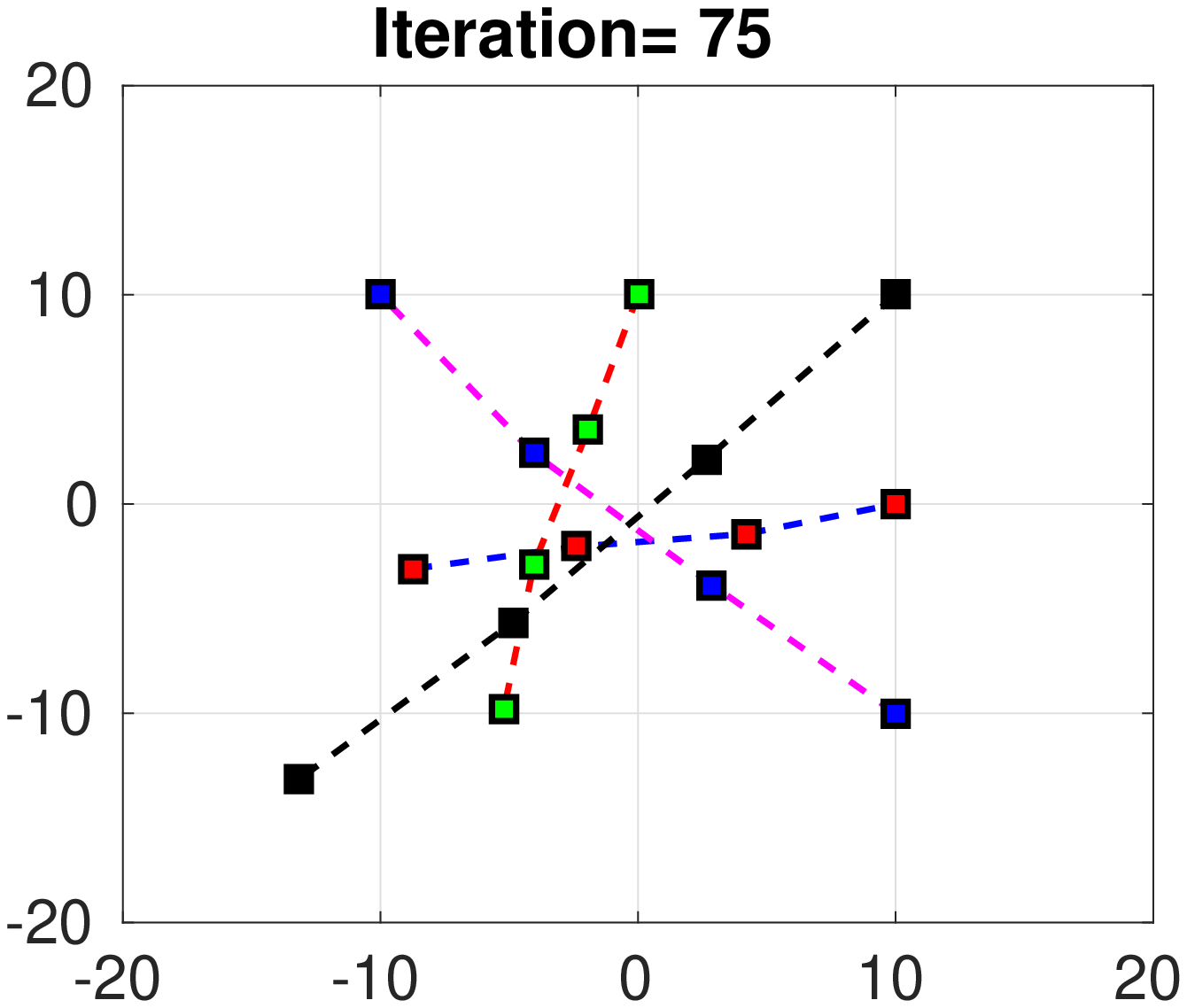}}\\
 \subfloat[]{\label{fig4:f150}\includegraphics[width=0.45\linewidth]{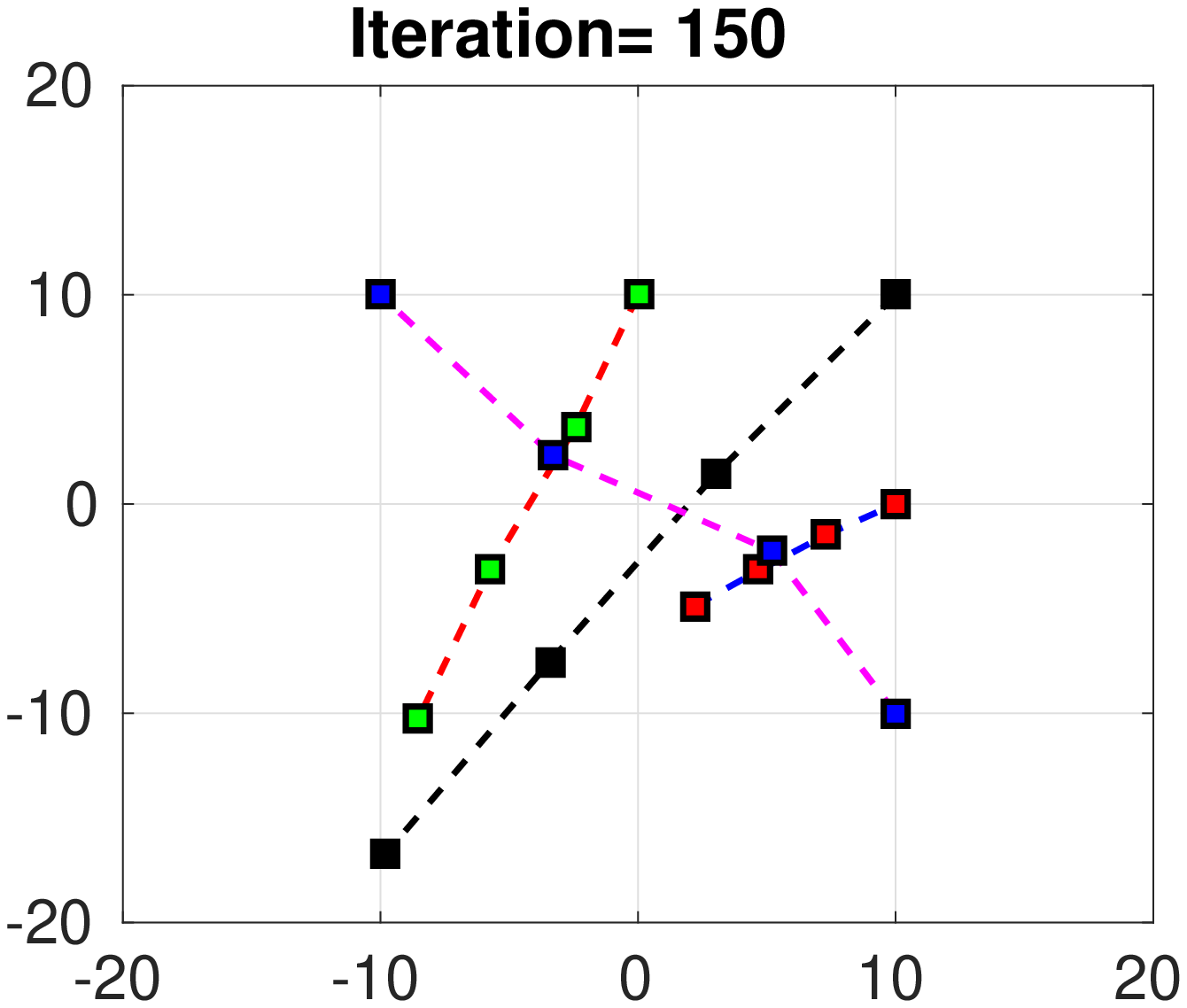}}\   
 \subfloat[]{\label{fig4:f200}\includegraphics[width=0.45\linewidth]{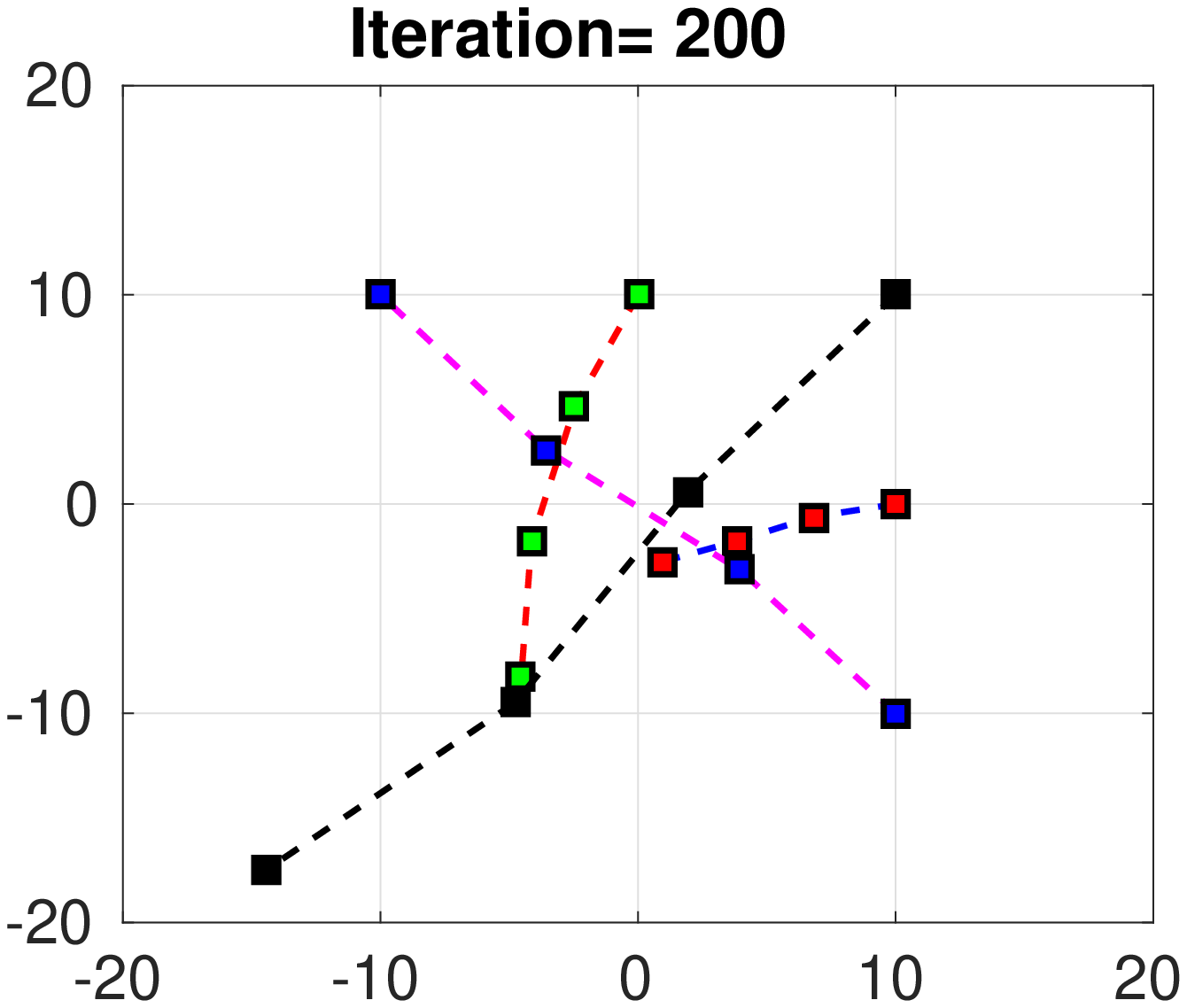}} 
 \caption{{Simulation plots: Configuration after (a) 1 Iteration (b) 75 Iterations (c) 150 Iterations (d) 200 Iterations}}
 \label{fig:4flows}
\end{figure}

\begin{figure}[ht!]
 \centering
 \includegraphics[width=\linewidth, height=0.5\linewidth]{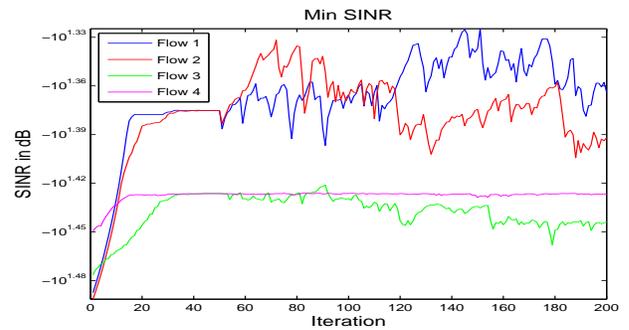}
 \caption{ Variation in the minimum SINR of each flow over time}
 \label{fig:sinr_var}
\end{figure}

So far, we have dealt with only two flow optimization problems with static endpoints, which are very simple compared to problems with higher complexity and multiple flows.
In order to test the performance of our algorithm in a more complex framework, we increased the complexity of the network by adding two extra flows with two robots assigned to each of the flow. 
The sender-receiver pair for flow 3 are initially located at $(10,10)$ and $(-10,-10)$, and for flow 4 are located at $(-10,10)$ and $(10,-10)$. 
We introduce random mobility pattern to Flow 1's transmitter, Flow 2's receiver and Flow 3's receiver.
The results of a simulation instance with this configuration is presented in Figure~\ref{fig:4flows}.
The figure demonstrates that our algorithm works well for a network with four flows with total 16 node and mobile endpoints.
Figure~\ref{fig:sinr_var} illustrates the convergences of the flow-wise minimum SINR values over time.\
It is clear from Figure~\ref{fig:sinr_var} that the minimum SINR of similar flows, i.e., Flow 1 and Flow 2, or Flow 3 and Flow 4, converges to the same values when the flow endpoints are static. Once the mobility is introduced, the SINRs change based on the new positions of the communication endpoints. 
We have also tested our algorithm for networks that are asymmetric and our algorithm performs equally well in those cases.\
However, we do not present those results in this paper to conform with the space limitation.

\section{Related Work}
\label{sec:related}
In this section, we provide a brief overview of the existing research works related to our field of interest.
Most of the significant research on robotic wireless router related topics are very recent.
Yan and Mostofi (\hspace{1sp}\cite{yan2010robotic,yan2012robotic}) are among the few researchers to work on the robotic router related problems.
They focused on robotic router placement optimization in order to maintain connectivity between an user and a base station.
This optimization problem is mainly focused on minimization of bit-error rate for two scenarios of multi-hop and diversity.
They also demonstrated that optimizations based on the Fiedler eigenvalue, instead of bit-error rate, result in a performance loss.
In these works, they used an extension of the channel modeling technique introduced in (\cite{mostofi2010estimation,malmirchegini2012spatial}).
However, they ignored the effect of interference in their model and focused on a single flow between a single receiver-transmitter pair.
Unlike these works, our proposed method is based on SINR,  which is more generalized than bit-error rate approach.
Also, we optimize multiple flows simultaneously, instead of focusing on a single flow.
Tekdas \emph{et al.}\cite{tekdas2010robotic}, also focused on similar problem and proposed two motion planning algorithms based on known user motion and unknown-random adversarial user motion, respectively.
Among other state-of-the-arts, the decentralized algorithm based on super-gradient and decentralized computation of Feidler eigenvector by DeGennaro and Jadbabaie\cite{degennaro2006decentralized} is mentionable.
Stump, Jadbabaie and Kumar \cite{stump2008connectivity} also developed a framework to control a team of robots based on two metrics: the Fiedler value of the weighted Laplacian matrix and the k-connectivity matrix.
However, Yan and Mostofi\cite{yan2010robotic} showed that Fiedler eigenvalue does not reflect the true reception quality, which is crucial in wireless networks.

Among other works, the DARPA LANdroids program \cite{mcclure2009darpa} is mentionable. 
Tactical communication enhancement in urban environments is the main goal of this program.
Towards this goal, they tried to develop pocket-sized intelligent autonomous robotic radio relay nodes, LANdroids, that are inexpensive.
LANdroids are used to mitigate the communications problems in urban settings, such as multipath effect, by acting like relay node into shadows, using autonomous movement and intelligent control algorithms. 
Dixon and Frew\cite{dixon2009maintaining} proposed a decentralized mobility controller based on maximizing the capacity of a local 3-node network in order to maximize the end to end capacity of the entire communication chain.
They used measurements of the local signal to noise ratio for this purpose.
A Disjunctive Programming Approach is presented in \cite{bezzo2011disjunctive}.  Among other works, the work of Vieira, Govindan
and Sukhatme~\cite{Vieira:2013kz} is mentionable.
In contrast, our proposed method is based on signal to interference and noise ratios and focuses on multiple flow optimization, which is more practical and generalized. 
  
In \cite{williams2013route}, Williams, Gasparri and Krishnamachari presented a hybrid architecture called INSPIRE, with two separate planes called Physical Control Plane (PCP) and Information Control Plane (ICP).
Their goal was to improve and optimize the network between multiple pair of senders and receivers using a group of robots and using ETX as a metric. They used ETX to determine the allocation of nodes among different flows, while the mobility framework is simply to place the robots evenly along the line segments joining the flow endpoints. 
Although our application contexts are the same, our mobility formulation as well as problem formulation are completely different. In our proposed model, the movement of the robots are directly controlled by the link qualities (more specifically, SINR) and, thus, is much practical.

\section{Conclusion}
\label{sec:concl}
In this paper, we have considered a problem of proper placement and control of mobile robotic nodes in order to optimize the performance of a wireless network.
We have devised an optimization function and based on that function, we have proposed a centralized and a distributed method of robotic node placement and control that maximizes our objective function.
Through a set of simulation experiments, we have demonstrated the performance and convergence of our algorithms.
However, due to space constraints, we have left out detailed description of some key portions of our algorithm such as the SINR modeling techniques as well as complexity and optimality analysis of our algorithm.
Therefore, our future direction will be to flesh out the details of an adaptive SINR model as well a thorough analysis of our algorithm.
This work is also a foundation of our future goal to develop an algorithm for adaptive node allocation and placement among different flows in order to handle various dynamic situation in the network such as flow addition or deletion and increase or decrease in flow demands. Another future direction would be a practical implementation of our algorithms on a real robotic network testbed.


{\footnotesize

\bibliographystyle{IEEEtran}
%

\bibliography{ref}
}
\end{document}